\documentclass[fleqn,10pt]{wlscirep}
\title{Machine Learned Learning Machines}

\author[1,2]{Leigh Sheneman}
\author[1,2,3,*]{Arend Hintze}
\affil[1]{Department of Computer Science and Engineering, Michigan State University, East Lansing, 48824, USA}
\affil[2]{BEACON-Center for the Study of Evolution in Action, Michigan State University, East Lansing, 48824, USA}
\affil[3]{Department of Integrative Biology, Michigan State University, East Lansing, 48824, USA}
\affil[*]{hintze@msu.edu}

\begin{abstract} 
There are two common approaches for optimizing the performance of a machine: genetic algorithms and machine learning. A genetic algorithm is applied over many generations whereas machine learning works by applying feedback until the system meets a performance threshold. These methods have been combined before in particular in artificial neural networks using an external objective feedback mechanism. We adapt this approach to Markov Brains, which are evolvable networks of probabilistic and deterministic logic gates. These MB so far could only adapt from one generation to the other, so we introduce \textit{feedback gates} which augment their ability to learn during their lifetime. We show that Markov Brains can incorporate these feedback gates in such a way that they do not rely on an external objective feedback signal, but instead can interpret the results of their action in order to create internal feedback in order to learn. This results in a more biologically accurate model of the evolution of learning, which will enable us to study the interplay between evolution and learning and could be another step towards autonomously learning machines.
\end{abstract}
\begin{document}

\flushbottom
\maketitle
%
%
\thispagestyle{empty}

\section*{Introduction}
Natural organisms not only have to fit their environment, but also have to adapt to changes in their environment which means that they have to be plastic. While plasticity occurs in many forms, here we focus on \emph{neural plasticity} which we define as an organisms ability to use ``experiences'' to improve later decisions and behavior. Being able to solve a T-maze repetitively, remembering where food is or avoiding places where predators have been spotted, and even learning another language are all cognitive abilities that require an organism to have neural plasticity. We will show how this neural plasticity can evolve in a computational model system. Neural plasticity allows natural organisms to learn due to reinforcement of their behavior\cite{hebb2005organization}. However, learning is tied to specific neural mechanisms- working memory (WM), short-term memory (STM) and long-term memory (LTM). While learning was initially perceived as a new ``factor'' in evolution\cite{baldwin1896new}, potentially even independent, it has since then been well integrated into the \textit{Modern Synthesis of Evolution}\cite{sznajder2012adaptive}. Evolution and learning can have a positive effect on each other\cite{hinton1987learning,fontanari1990effect}, however, this is not necessarily always the case\cite{santos2015phenotypic}. This has several implications: At first evolution started with organisms that could not adapt during their lifetime, which means that they had no neural plasticity. The only feedback that the evolutionary process receives is differential birth and death. As a consequence, learning will only evolve if it can increase the number of viable offspring, and it can only do so, if there is a signal that predictably indicates a fitness advantage\cite{dunlap2016reliability}. 

Organisms receive many signals from their environment which have to be filtered and interpreted. Irrelevant signals should be ignored while others require adaptive responses. This can be done through instincts or reflexes in cases where a fixed responses are necessary. In other cases information has to be stored and integrated in order to inform later decisions, which requires memory and learning. To distinguish between actions that lead to advantageous results and those who are disadvantageous organisms need positive or negative feedback. However, none of the signals organisms receive are inherently ``good'' or ``bad''; even a signal as simple as food requires interpretation. The consumption of food has to trigger a positive feedback within the organism in order to function as a reward. The machinery that triggers the feedback is an evolved mechanism and is often adaptive to the environment. If food would be a global positive feedback signal, it would reinforce indiscriminate food consumption. Organisms would not be able to avoid food or store it for later, but instead eat constantly.

Another important detail we have to consider is the difference between learning and memory. While memory is information about the past, learning is the process that takes a sensorial percept and, typically by reinforcement, retains that information for later use. Specifically, sensor information is stored in working memory (WM)\cite{Ma:2014, Nadel:2010}. Imagine this as the flurry of action potentials that goes through the brain defining its current state. Information that a living organism needs to store for a moment is believed to reside in STM\cite{Nadel:2010, Kandel:2014}, but how information transforms from WM to STM is not fully understood\cite{Kandel:2014, Ma:2014, Squire:2011}. Natural systems use their LTM if they want to keep information for longer. Presumably, information from STM becomes reinforced and thus forms LTM, this is sometimes referred to as consolidation\cite{Nadel:2010, Kandel:2014,Abraham:2005}. The reinforcement process takes time and therefore is less immediate than STM. In addition, memories can be episodic or semantic\cite{McKenzie:2011,Kandel:2014,Nadel:2010} and can later be retrieved to influence current decisions. While information in the working memory can be used to influence decisions, it does not change the cognitive substrate, long term potentiation (or other neural processes) on the other hand use this information to change the neural substrate, by presumably forming or modifying connections. 

In summary, if we want to model the evolution of learning in natural organisms properly, we need to take the following statements serious:
\begin{itemize}
    \item evolution happens over generations, while learning happens during the lifetime of an organism 
    \item evolution is based on differential birth and death (selection) and learning evolved to increase the number of viable offspring and/or to avoid death
    \item organisms do not receive an objective  
    ``positive'' or ``negative'' signal, but instead evolved mechanisms to sense and interpret the world so that they can tell what actions were positive and which ones were not
    \item memory is information about the past, which can be transient
    \item information in the WM does not change the cognitive machinery, while learning changes the substrate to retain information for longer, turning transient into permanent information
\end{itemize}

\subsection{Machine Learning}
Computer science and engineering are typically not concerned with biological accuracy but more with scalability, speed, and required resources. Therefore, the field of machine learning is much more of a conglomerate of different methods, which straddle the distinct concepts we laid out above. Machine learning includes methods such as data mining, clustering, classification, and evolutionary computation\cite{russell2009}. Typically, these methods try to find a solution to a specific problem. If we provide an explicit reference or example class we refer to supervised learning since the answer is known and the fitness function quantifies the difference between the provided solution and the ones the machine generates. For unsupervised learning we provide a fitness function that measures how well a machine or algorithm performs without the need to know the solution in advance. Genetic algorithms (GAs), which are a form of evolutionary search, work in supervised or unsupervised contexts, whereas learning algorithms are typically supervised. A special class are learning to learn algorithms, which improve their learning ability while adapting to a problem\cite{schmidhuber1987evolutionary} but do not necessarily apply evolutionary principles.

Genetic algorithms clearly optimize from one generation to the other, while learning algorithms on the other hand could be understood as lifetime learning. $Q$-learning\cite{watkins1989} optimizes a Markov Decision Processes by changing probabilities when a reward is applied. Typically, delayed rewards are a problem, which deep-$Q$ learning and memory replay try to overcome\cite{mnih2013playing}. Artificial neural networks can be trained by using back propagation\cite{Schmidhuber:2015, Zhang:2012, LeCun:2015}, the Baum-Welch algorithm\cite{Kaleh:1994,Baggenstoss:2001}, or gradient decent\cite{williams1992simple,kaelbling1996reinforcement} which happens episodically, but on an individual level and not to a population that experiences generations. Multiplicative weights algorithm strengthens or weakens connections in a neural network-based on the consensus of a pool of experts\cite{Arora:2012, Freund:1999}, again on an individual.

At the same time, memory and learning are often treated interchangeably. Recurrent artificial neural networks can store information in their recurrent nodes (or layer) without changing its weights, which would be analagous to WM. Similarly, the system we use, Markov Brains (MB), can form representations about their environment and stores this information in hidden states (WM), again transiently without changing its computational structure\cite{Marstaller:2013}. Changes to the weights of an ANN, or the probabilities of a Markov process or POMDP\cite{kaelbling1998planning} for example, reflect much better learning, since those changes are not transient, and change all future computations executed by the system.

\begin{figure}[ht]
\centering
\includegraphics[width=6.5in]{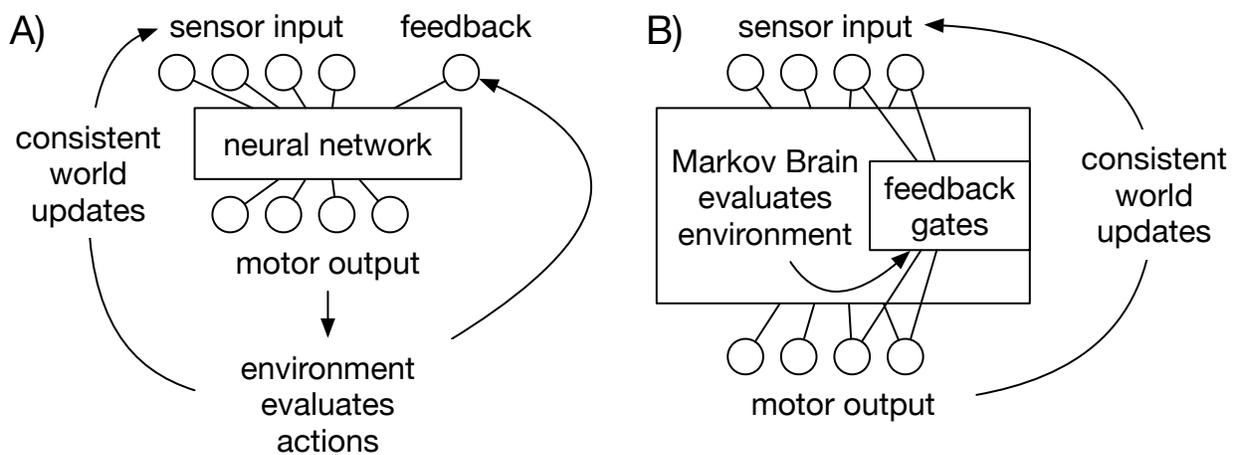}
\caption{Comparison of two different approaches to feedback generation. Traditional methods have the environment evaluate the actions of a neural network (or similar system) and then generate feedback which is provided as an external signal, very much like yet another input (panel A). Our approach (panel B) integrates the feedback generation as well as the feedback gates into the Markov Brain. This way, only the world provides consistent information about itself which can be evaluated so that feedback generation does not need to be offloaded, but becomes part of the entire machine. The entire connectivity as well as feedback gates are integrated is part of what needs to evolve.
\label{fig:overviewIvO}} 
\end{figure}

We also find a wide range of evolvable neural network systems\cite{sims1994evolving,stanley2002evolving,gauci2010autonomous} (among many others) which change from generation to generation and allowing for memory to form by using recurrent connections. Alternatively, other forms of evolving systems interact and use additional forms of memory\cite{spector2002genetic,greve2015evolving}. In order to evolve and learn, other systems allow the topology and or weights of the neural network to change while also allowing weight changes during their lifetime\cite{yao1999evolving,stanley2003evolving,gomez2005evolving,urzelai2006evolution,soltoggio2008evolutionary,luderscontinual,tonelli2011relationships,risi2012unified,coleman2012evolving,greve2015evolving,greve2016evolving}. Presenting objective feedback in order to adapt these systems during their lifetime allowed their performance to improve. As a consequence, the machinery that interprets the environment in order to create feedback was of no concern, but, as stated above, natural organisms need to evolve that machinery as well in order to learn. We think it is quite possible to change these systems to not rely only on external feedback. Instead, they could create the feedback signal as part of their output themselves. However, none of the systems mentioned above is an evolvable MB (for a comparison see Figure \ref{fig:overviewIvO} A vs. B). 

In our approach we use MB\cite{Edlund:2011}, which are networks of deterministic and probabilistic logic gates, encoded in such a way that Darwinian evolution can easily improve them. MB have been proven to be a useful tool to study animal behavior\cite{Olson:2013,Hintze:14}, neural correlates\cite{Joshi:2013,Marstaller:2013,Albantakis:2014},evolutionary dynamics\cite{Schossau2016}, decision making\cite{kvam2015,Kvam2017}, and can even be used as a machine learning tool\cite{Chapman:2013,chapman2017evolution}. One can think of these MBs as artificial neural networks (ANN)\cite{russell2005ai} with an arbitrary topology that uses Boolean logic instead of logistic functions. Through sensors these networks receive information about their environment as zeros or ones, perform computations and typically act upon their environment through their outputs. We commonly refer to MBs that are embodied and through that embodiment\cite{clark1998being} interact with their environment as agents (others use the term animat which is synonymous). MB use hidden states to store information about the past similar to recurrent nodes in an ANN. The state of these hidden nodes has to be actively maintained, making the information volatile. The information in the hidden states can be used to perform computations and also functions as memory. This form of memory, however, resembles WM or STM much better than LTM due its volatile nature. The entire structure of a MB to this point would be encoded by the genome and would not change over the lifetime of the agent. Here we introduce what we call feedback gates, which allow MBs to use internal feedback to store information by changing their probabilistic logic gates (see Methods for a detailed description of feedback gates). Like other systems these updates do not change the topological structure of the node network but rather the probabilities within the gates, similar to how learning in ANN is achieved through weight changes. However, feedback is not an objective signal coming from the environment, but must be generated as part of the evolved controller. The feedback gates only receive internally generated feedback to change their behavior. This linkage between inputs, evaluation of the environment to generate feedback, how feedback gates receive this information, and how everything controls the actions of the agent evolves over time (see Figure \ref{fig:overviewIvO} B). 

Here we introduce feedback gates allowing MBs to change their internal structure which is akin to learning and forming long-term memories during the lifetime of an organism. The closest similarity can be found in $Q$-learning or $Q$-deep learning, which changes the probabilities within a Markov Decision process even if rewards are delayed\cite{russell2009}. We will show that these feedback gates function as expected and allow agents to evolve the ability to decipher sensor inputs so that they can autonomously learn to navigate withing a complex environment.

\section*{Results}
Natural organisms have to learn many things over their lifetime including how to control their body. Here the environment used to evolve agents resembles this problem. Agents have to learn to use their body in order to navigate properly. The environment is a 2D lattice (64x64 tile wide) where a single tile is randomly selected as the goal an agent must reach. The lattice is surrounded by a wall so agents can not escape the boundary, and $\frac{1}{7}$ of the lattice is filled with additional walls to make navigation harder. From the goal the Dijkstra's path is computed so that each tile in the lattice can now indicate which of its neighboring tiles is the next closest to the goal. In cases where two neighbor tiles might have the same distance, one of these tiles is randomly selected as the next closest. For ease of illustration we can now say that a tile has an arrow pointing towards the tile that should be visited next to reach the goal in the shortest number of tiles. 

The agent, controlled by a MB, is randomly placed on a tile that is 32 tiles away from the goal and facing in a random direction (north, west, south, or east). Agents can see the arrow of the tile they are standing on. The direction indicated by the tile is relative to the that of the agent, so that a tile indicating north, will only be perceived as a forward facing arrow if the agent also faces north. The agent has four binary sensors that are used to indicate in which relative direction the agent should go to reach the goal.

The agent can move over the lattice by either turning 90 degrees to the left or right, or by moving forward. So far, in order to navigate perfectly, the agent would simply need to move forward when seeing a forward facing arrow, or turn accordingly. 
Instead of allowing the agent to directly pick a movement, it can choose one of four intermediate options (A,B,C,or D) at any given update. At the birth of an agent, these four possible options are mapped to four possible actions: move forward, turn left, turn right, do nothing. As a result, the complexity of the task increases when the agent has to learn which of the 24 possible option-to-action maps currently applies to navigate the environment properly. The agent is not given any direct feedback about its actions; a mechanism must evolve to discern the current mapping and this is rather difficult. 

In prior experiments\cite{Edlund:2011,Marstaller:2013,Albantakis:2014,Chapman:2013,Hintze:14,Joshi:2013,kvam2015,Schossau2016,Olson:2013}, MBs were made from deterministic or probabilistic logic gates that use a logic table to determine the output given a particular input. Deterministic gates have one possible output for each input, while probabilistic gates use a linear vector of probabilities to determine the likelihood for any of the possible outputs to occur.
To enable agents to form LTM and learn during their lifetime we introduce a new type of gate: feedback gate. These gates are different from other probabilistic gates, in that they can change their probability distribution during their lifetime based on feedback (for a detailed description, see below). This allows for permanent changes which are akin to LTM. While Markov Brains could already retain information by using hidden states, now they can also change ``physically''. With this final requirement, MBs must evolve to integrate these new gates into their network of other gates and find a way to supply feedback appropriately.

\subsection*{Feedback Gate Usage}
To test if the newly introduced feedback gates help evolution and increase performance, we compare three different evolutionary experimental conditions. Agents were evolved over 500,000 generations that could use only deterministic logic gates, deterministic and probabilistic logic gates, or all three types of gates--deterministic, probabilistic, and feedback gates to solve the task. 

When analyzing the line of descent (LOD; see materials and methods), we find a strong difference in performance across the three evolutionary conditions (see supplementary information Figure 1). None of the 300 agents that were evolved using deterministic together with probabilistic logic gates were capable of reaching the goal in any of the 24 mappings. The agents that were allowed to use only deterministic logic gates failed to reach the goal in 225 of the 300 experiments, but the remaining 75 agents never reached the goal more than five times. Agents allowed to use all gates including feedback gates only failed to reach the goal in 75 experiments and in the remaining 225 experiments they reached the goal on average 5 times, with the best performer reaching the goal on average 9 times. 

It is not entirely surprising to us that agents using only probabilistic gates struggle in this task, because agents using probabilistic gates generally evolve slower, which might explain the effect. However, to our surprise, we found a couple of agents who were only allowed to use deterministic gates that evolved to solve the task at least a couple of times. In the group that could use all three types of gates, we found 3 agents that could reach the goal on average 5 times using only probabilistic gates, as opposed to the 9 times agent with feedback gates could reach the goal on average. This shows two things: the task can be solved using only the gate inputs (i.e. WM) and providing agents with feedback gates during evolution allows them to reach the goal more often.  This is an important control because from a computational point of view, there is no qualitative difference between WM and LTM as both methods allow for recall of the past.

\begin{figure}[ht]
\centering
\includegraphics[width=6.5in]{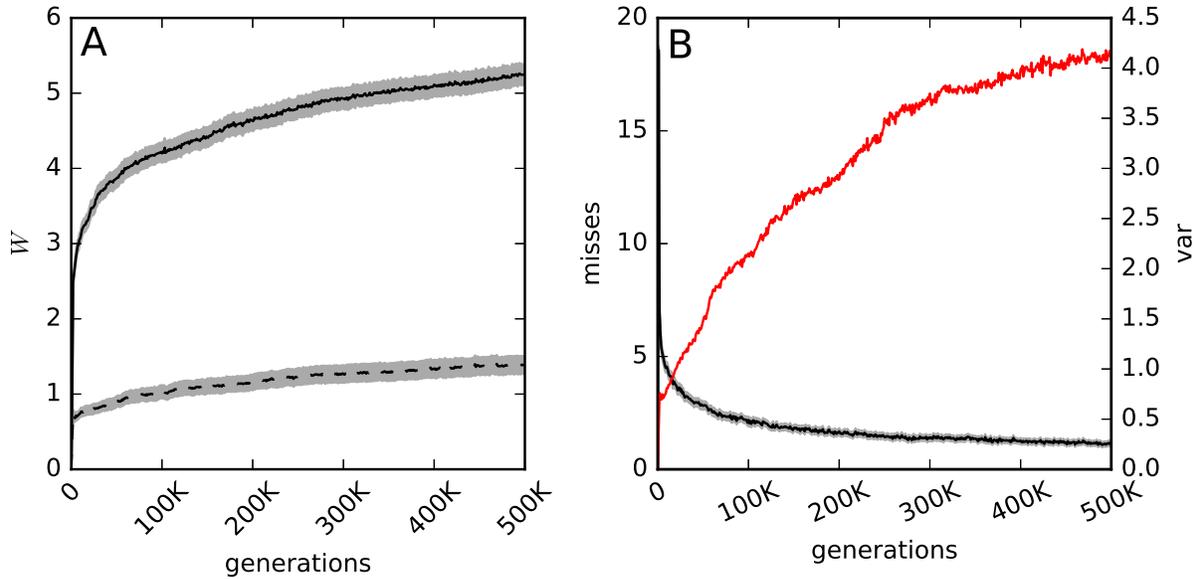}
\caption{Performance over evolutionary time. Panel A solid line shows for all 300 replicate experiments how often the goal was reached ($W$) across all 24 possible environments for agents on the line of descent. The dotted line is the same average performance for the same agents when their feedback mechanism was disabled. Underlying gray-shade indicates the standard error. Panel B shows as a black line how often on average the 300 replicate agents on the line of descent could not reach the goal a single time in any of the 24 possible environments. In red the variance in performance on the line of descent as an average over all 300 replicate experiments.
\label{fig:fitnessMissesVar}} 
\end{figure}

Populations allowed to use feedback gates quickly evolve the ability to reach the goal in any of the 24 possible environments. The variance of their performance supports the same idea, that agents do not become better by just performing well in one environment, but instead evolve the general ability to learn the mapping each environment presents (See Figure \ref{fig:fitnessMissesVar} panel B). 

Now that we have shown that the agents with feedback gates are capable of evolving a solution to navigate in this environment, we have to ask if they actually utilize the feedback gates. For that, all agents on the LOD were tested again but their feedback gates were kept from changing their probability tables. Comparing these results with the agents performance when using the feedback gates regularly reveals that the agents rely heavily on their feedback gates (see \ref{fig:fitnessMissesVar} panel A). 

\subsection*{Feedback gates change over time}
\begin{figure}[ht]
\centering
\includegraphics[width=3.5in]{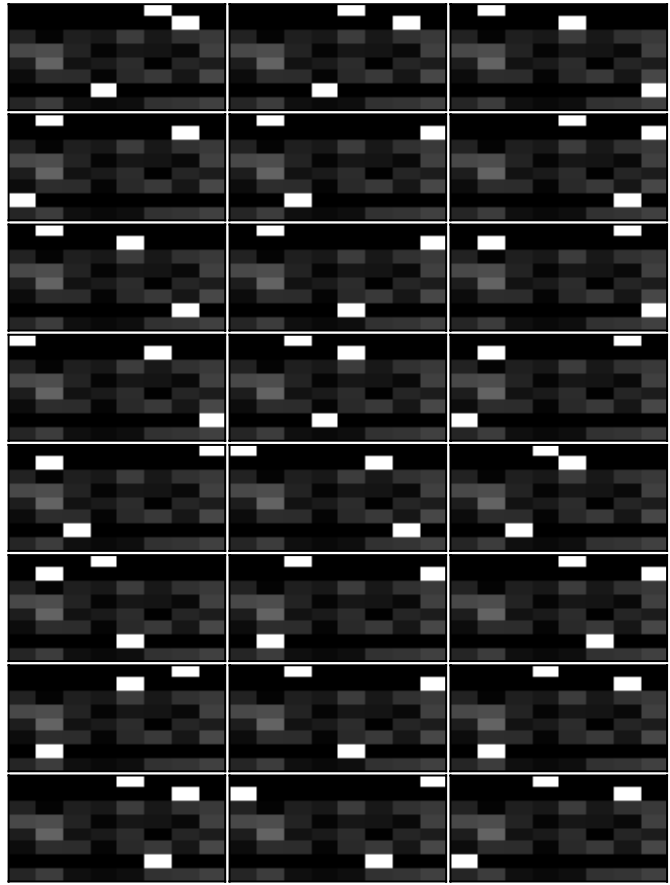}
\caption{Probability tables of a feedback gate after it learned each of the 24 possible mappings. Each of the 24 gray scale images corresponds to a feedback table adapted during the lifetime of an agent to a different mapping. The darker the color, the lower the probability, observe that rows have to sum to 1.0. Some rows, apparently those of input combinations that were never experienced, remain unadapted, while rows one, two, and seven of each table converge to a single high value surrounded by low probabilities.
\label{fig:mappingAgent3}}

\end{figure}

We find that the probability tables modified by feedback become specifically adapted to each of the 24 possible mappings the agents get tested in. See Figure \ref{fig:mappingAgent3} as an example of the best performing agent using only one feedback gate. Some rows in the probability tables converge to having a single high value that is specific to the environment the agent experienced (for more details see supplementary information Figures 2-3). This shows, that indeed feedback gates become specifically adapted to the environment the agent experiences. It also indicates, that agents change their computational machinery according to their environment and do not rely solely on WM to perform their task.

\begin{figure}[ht]
\centering
\includegraphics[width=4.5in]{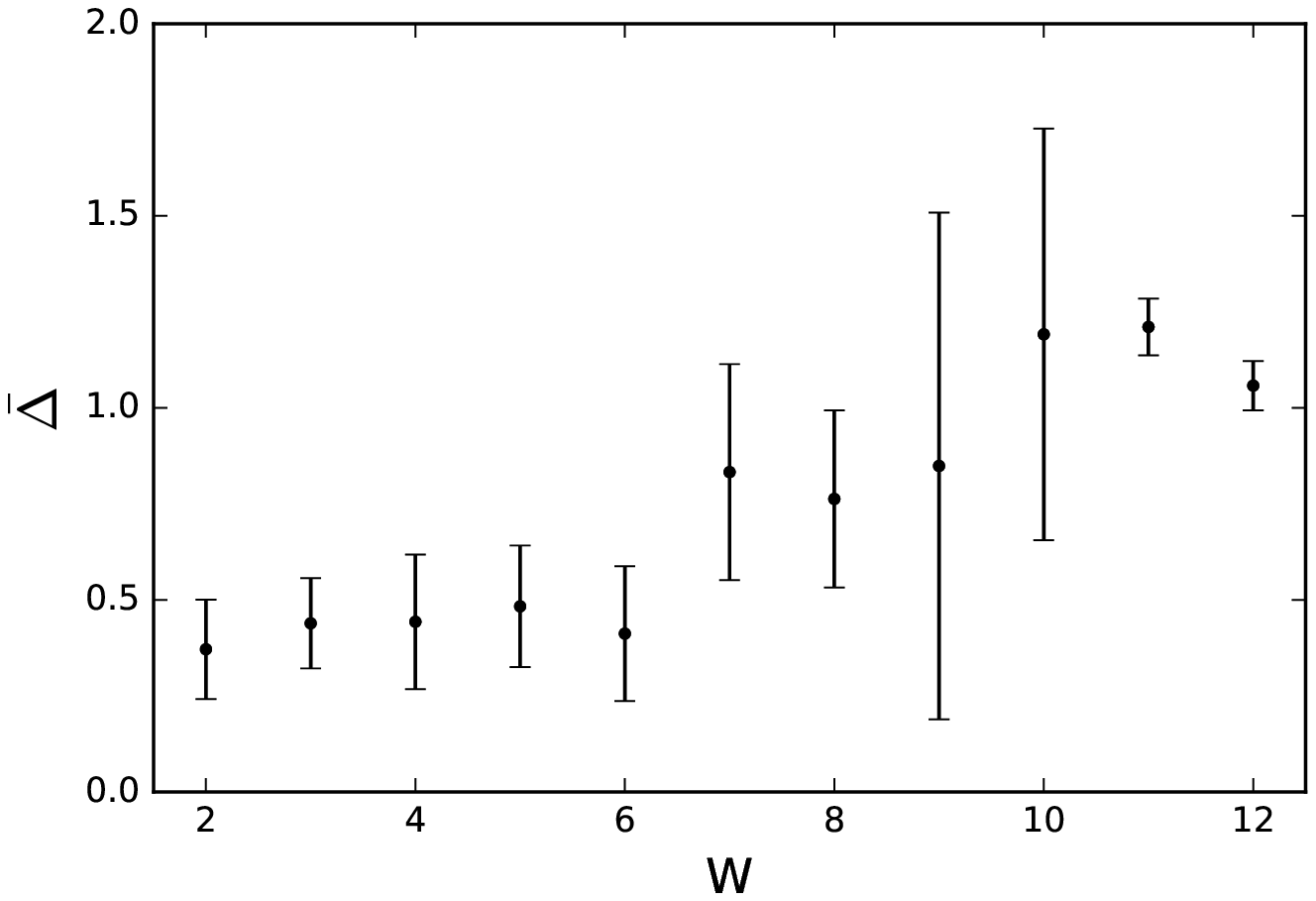}
\caption{Change in mutual information of feedback gates for different levels of performance. The change in performance ($\bar{\Delta}$) is shown on the y-axis for different performances (W). The performance for all 300 final organisms using all types of gates was measured and put into ten bins, ranging from the lowest performance 2 to the highest performance of 12, with a bin size of 1. Error-bars indicate the variance.\label{fig:DeltaMutualOverPools}}
\end{figure}

The change to the feedback gates' probability tables can be quantified by measuring the mutual information each table conveys at birth and after the agent completes their task. We find that the mutual information is generally lower at birth (${\sim}0.25$) and higher at the end of the task (${\sim}0.8$), signifying that the agents have more information about the environment at death than they did at birth, as expected. We then compute the difference between both measurements, ($\bar{\Delta}$), which quantifies the increase of mutual information over the lifetime of the agent. When discretizing these values for different levels of performance, we find a strong correlation ($r=0.922$) between performance and increase in mutual information (see Figure: \ref{fig:DeltaMutualOverPools}). This shows that agents who perform better increase information stored in their feedback gates over their lifetime.

\subsection*{Differences between agents using feedback gates and those who do not}
We wanted to test if feedback gates improve the agents ability to learn. Those agents that evolved to use feedback gates generally perform very well in the environment, however, we also find other agents that only use deterministic gates and still have an acceptable performance. This either suggests that feedback gates are not necessary, or that they do not provide enough of an selective advantage to be used every time. It is also possible that there is more than one algorithmic solution to perform well in this navigation task. All of these points suggest that a more thorough analysis and comparison of the independently evolved agents is necessary. 

\begin{figure}[ht]
\centering
\includegraphics[width=2.5in]{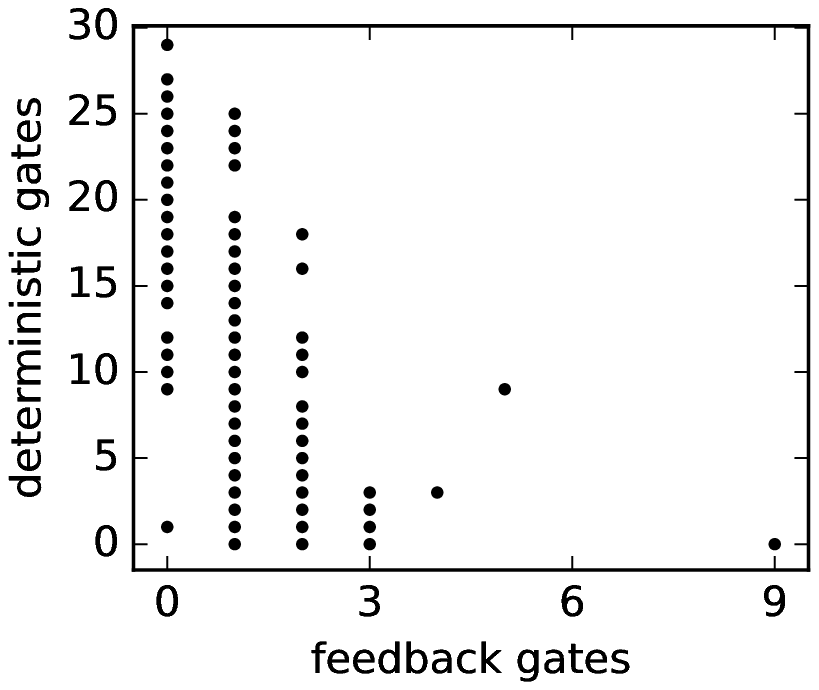}
\caption{Correlation of number of feedback gates to deterministic gates. Each dot represents the number of feedback gates last agent on the line of descent has versus the number of deterministic gates it has in 300 replicates. The more feedback gates an agent has the less deterministic gates it evolves; the feedback gates allow agents to decrease brain size while still solving the task.\label{fig:FvsD}}

\end{figure}

We find that agents that do not use feedback gates require a much greater number of logic gates than those who do (see Figure \ref{fig:FvsD}). This seems intuitive, since feedback gates can store information in their probability tables, whereas agents that do not use them, need to store all information in their WM. This suggests that there might be a difference in the evolved strategy between those agents that use feedback gates and those who do not.When observing the behavior of the differently evolved agents, we find that there are two types of strategies (see supplementary information Figures 4-5 for details). Agents that evolved brains that without feedback gates use a simple heuristic that makes them repeat their last action when the arrow they stand on points into the direction they are standing on. Otherwise, they start rotating until they move off a tile, which often results in them standing again on a tile that points into the direction they are facing, which makes them repeat the last action. 

\begin{figure}[ht]
\centering
\includegraphics[width=4.5in]{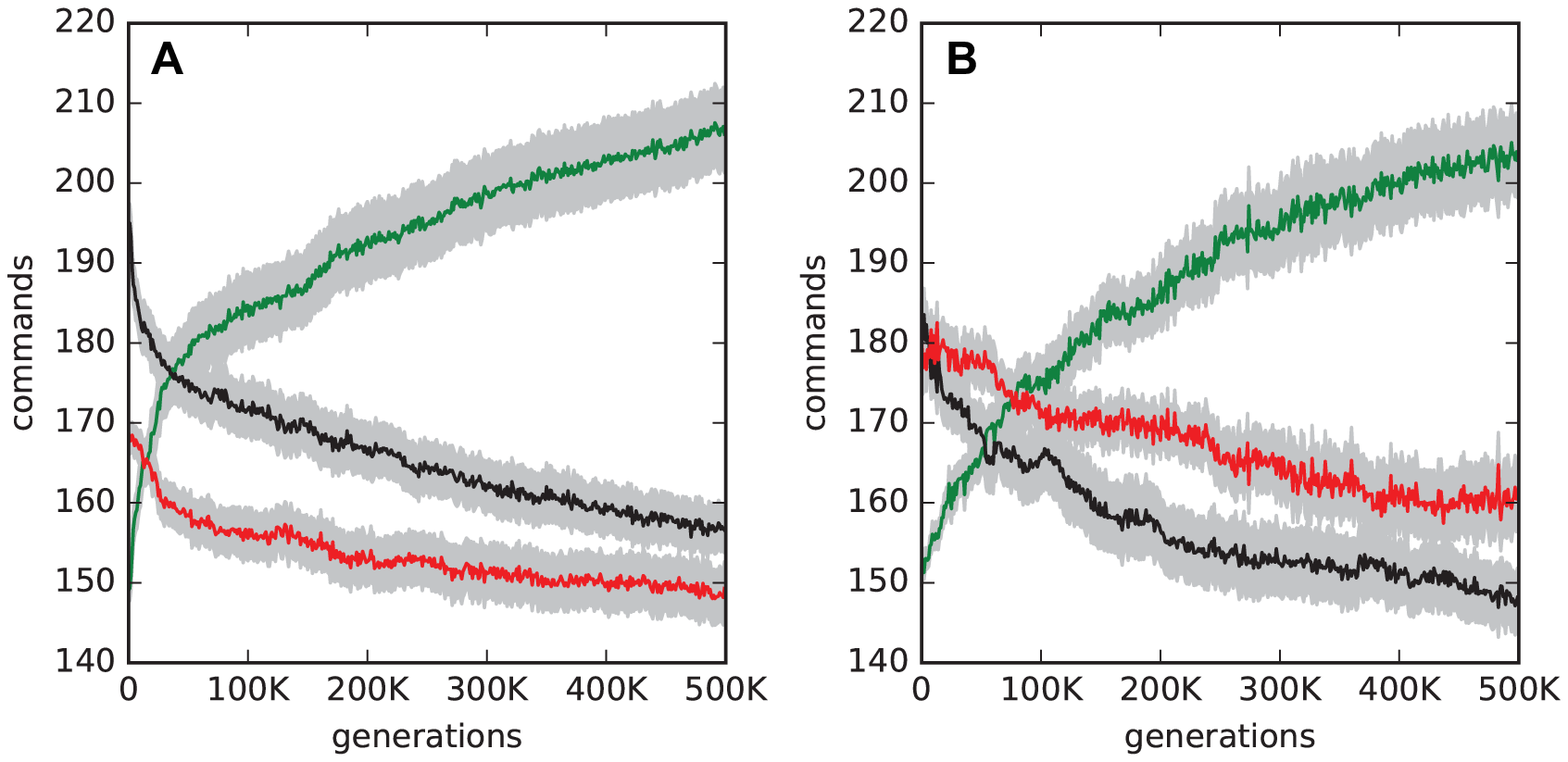}
\caption{Different actions performed while navigating over evolutionary time. Both panels show the average use of forward (green), do nothing (red), and turn (black) commands over generations. Gray background indicates the standard error. Panel A for those 244 agents that do use feedback gates, Panel B for the remaining 56 independently evolved agents that do not use feedback gates. Agents on the LOD were analyzed.\label{fig:commandUsage}}
\end{figure}

Agents that evolved to use feedback gates appear to actually behave as if they learn how to turn properly. They make several mistakes in the beginning, but after some learning period perform flawlessly. Of the 300 replicate evolutionary experiments where agents were allowed to use all types of logic gates, 56 did not evolve using feedback gates. Comparing the actions those 56 agents take, with the remaining 244 which do use feedback gates, we first find that as expected the group using feedback gates reached the goal more often on average ($5.33$ times, versus $4.89$ times for those agents not using feedback gates) which suggests a difference in behavior. The usage of actions is also drastically different even during evolution (see Figure \ref{fig:commandUsage}). 
Agents using feedback gates reduce the instances where they do nothing, minimize turns and maximize moving forward. Agents not using feedback gates are less efficient because they rely on forward movements while minimizing times where they do nothing and turns. In conjunction with the observations made before, we conclude that indeed agents not using feedback gates, use some form of heuristic with a minimal amount of memory, while agents using feedback gates learn to navigate the environment properly. 

\section*{Discussion}
We use Markov Brains which could evolve to solve various complex tasks and even form memories which represent their environment\cite{Marstaller:2013}. However, these MBs use binary hidden states to store information about the environment similar to WM or STM, but lacked a feedback mechanism that allowed them to change their internal structure, very much like long term potentiation would lead to LTM. Other systems like artificial neural networks already allowed for a similar process, were weights could be adapted due to feedback. We augmented Markov Brains with so called feedback gates that use the multiplicative weight update algorithm to change their probabilities given positive or negative feedback. 

It is important to note, that this feedback is not directly provided by the environment the agents but must be internally generated by the MB. In addition, MB need to integrate these feedback gates into their cognitive machinery during evolution. We showed that we can successfully evolve Markov Brains to use feedback gates and that it is possible that internal feedback is generated properly and that feedback gates change their internal probabilities as expected. However, we also find competing strategies, which instead of evolving to learn deploy some form of heuristic. In the future it might be interesting to study under which evolutionary conditions either heuristics or learning strategies evolve (similar to Kvam 2017). We used a biological inspired task and we see the future application of this technology in the domain of computational modeling to study how intelligence evolved in natural systems, to eventually use neuroevolution as the means to bring about general artificial intelligence. Using MBs augmented with feedback will probably not be competitive with other supervised learning techniques applied to classification problems for example, but it remains an interesting question: How would typical machine learning tools perform if challenged with the task presented here?

One problem encountered by systems that receive external feedback is ``catastrophic forgetting''\cite{french1999catastrophic,ellefsen2015neural}. When the task changes, feedback will cause the system to adapt to their new objective but they forget or unlearn former capabilities. Requiring adaptive systems to interpret the environment in order to generate internal feedback, might be a way to overcome this problem, assuming that the fitness function or evolutionary environment not only changes as well, but rewards organisms who do not suffer from ``catastrophic forgetting''.

We now have the ability to evolve machines inspired by natural organisms that can learn without an objective external signal. This gives us a tool to study the evolution of learning using a computational model system instead of having to use natural organisms. It will also allow us to study the interplay between evolution and learning (aka Baldwin effect) and explore under which circumstances evolution benefits from learning and when it does not, and we propose to use this model to study these questions in the future. Another dimension we will investigate in the future is the ability of MBs to change their connections due to feedback, not just the probabilities within their gates.

As stated before, combining evolution with learning is not a new idea. We think that it is in principle very easy for other systems to internalize the feedback. For example, it should be easy to evolve an artificial neural network to first interpret the environment, and then use this information to apply feedback on itself. However, we need to ask under which circumstances this is necessary. By providing training classes for supervised learning situations we can already create (or deep learn) machines that can learn to classify these classes.  In addition, we often find these classifiers exceed human performance\cite{ciregan2012multi,silver2016mastering}. If we are incapable of providing examples of correctly classified data, we use unsupervised learning methods, and only need to provide a fitness function that quantifies performance. But we know that fitness functions can be deceptive, and designing them is sometimes more of an art than a science. When interacting with future AI systems we should find a different way to specify what we need them to do. Ideally they should autonomously explore the environment and learn everything there is to know without human intervention - nobody tells us humans what to research and explore, evolution primed us to pursue this autonomously. The work presented here is one step into this direction, and will allow us to study evolution of learning in a biological context as well as explore how we can evolve machines to autonomously learn.

\section*{Methods}
Markov Brains are networks of probabilistic and deterministic logic gates encoded by a genome. The genome contains genes and each gene specifies one logic gate, the logic it performs and how it is connected to sensors and motors and to other gates\cite{Edlund:2011,Marstaller:2013,Schossau2016}.  A new type of gate, the feedback gate, has been added to the Markov Brain framework (https://github.com/lsheneman/PROJECTREPO),and this framework has been used to run all the evolutionary experiments. The Markov Brain framework has since been updated to MABE\cite{MABE2016}. See below for a detailed description of each component: 

\subsection*{Environment}
The environment the agents had to navigate was a 2D spatial grid of 64x64 squares. Squares were either empty or contained a solid block that could not be traversed. The environment was surrounded by those solid blocks to prevent the navigating agent to leave that space. At the beginning of each agent evaluation a new environment was generated and $\frac{1}{7}$ of the squares were randomly filled with a solid block. The randomness of the environments maintains a complex maze-like structure across environments, but no two agents saw the exact same environment.

A target was randomly placed in the environment, and Dijkstra's algorithm was used to compute the distance from all empty squares to the target block. These distances were used to label each empty block so that it had an arrow facing to the next closest block to the target. When there was ambiguity (two adjacent blocks had the same distance) a random block of the set of closest blocks was chosen. At birth agents were randomly placed in a square that had a Dijkstra's number of 32 and face a random direction (up, right, down, or left). Due to the random placement of blocks it was possible that the goal was blocked so that there was no tile that is 32 tiles away, in which case a new environment was created, which happened only very rarely.

Agents were then allowed to move around the environment for 512 updates. If they were able to reach the target, a new random start orientation and location with a Dijkstra's number of 32 was selected. Agents used two binary outputs from the MB to indicate their actions-- 00, 01, 10, or 11. 
Each output was translated using a mapping function to one of four possible actions- move forward, do nothing, turn left, or turn right. This resulted in 24 different ways to map the four possible outputs of the MB to the four possible actions that moved the agent. The input sensors gave information about the label of the tile the agent was standing on. Observe that the agent itself had an orientation and the label was interpreted relative to the direction the agent faced. There were four possible arrows the agent could see-- forward, right, backward, or left-- and were encoded as four binary inputs, one for each possible direction.
Beyond the four input and two outputs nodes, agents could use 10 hidden nodes to connect their logic gates. Performance (or fitness) was calculated by exposing the agent to all 24 mappings and testing how often it reached the goal within the 512 updates it was allowed to explore the world. At every update agents were rewarded proportional to their distance to the goal ($d$), and received a bonus ($b$) every time they reached the goal, thus the fitness function becomes:
\begin{equation}
W=\prod_{n=0}^{n<24}((\sum_{t=0}^{t<512}\frac{1}{1+d})+b) \label{fitnessFunction}
\end{equation}

\subsection*{Selection}
After all agents in a population were tested on all 24 action-to-behavior mappings at each generation, the next generation was selected using tournament selection where individuals are randomly selected and the one with the best fitness transmits offspring into the next generation\cite{blickle1996comparison}. The tournament size was set to five.

\subsection*{Mutation}
Genomes for organisms in the first generation were generated randomly with a length of $5000$ and $12$ start codons were inserted that coded for deterministic, probabilistic, and feedback gates. Each organism propagated into the next generation inherited the genome of its ancestor. The genome had at least $1,000$ and at most $20,000$ sites. Each site had a $0.003$ chance to be mutated. If the genome had not reached its maximum size stretches of a randomly selected length between $128$ and $512$ nucleotides got copied and inserted at random locations with a $0.02$ likelihood. This allowed for gene duplications. If the genome was above $1000$ nucleotides, there was a $0.02$ chance for a stretch of a randomly selected length between $128$ and $255$ nucleotides to be deleted at a random location. 

\subsection*{Feedback Gates}
At every update of a probabilistic gate, an input $i$ resulted in a specific output $o$. To encode the mapping between all possible inputs and outputs of a gate we used a probability matrix $P$. Each element of this matrix $P_{io}$ defined the probability that given the input $i$ the output $o$ occurred. Observe that for each $i$ the sum over all $o$ must be $1.0$ to define a probability:
\begin{equation}
1.0=\sum_{o=0}^{O}{P_{io}} \label{equ:rowsSumToOneFunction}
\end{equation}
where $O$ defines the maximum number of possible outputs of each gate.

A feedback gate uses this mechanism to determine its output at every given update. However, at each update we consider the probability $P_{io}$ that resulted in the gates output to be causally responsible. If that input-output mapping for that update was appropriate then in future updates that probability should be higher. If the response of the gate at that update had negative consequences, then the probability should be lower. As explained above, the sum over all probabilities for a given input must sum to one. Therefore, a single probability can not change independently. If a probability is changed, the other probabilities are normalized so that equation \ref{equ:rowsSumToOneFunction} remains true.

But where is the feedback coming from that defines whether or not the action of that gate was negative or positive? Feedback gates posses two more inputs, one for a positive, and one for a negative signal. These inputs can come from any of the nodes the Markov Brain has at its disposal, and are genetically encoded. Therefore, the feedback can be a sensor, or any output of another gate. Receiving a $0$ from another gate to either of the two positive or negative feedback inputs has no effect, whereas reading a $1$ triggers the feedback. 

In the simplest case of feedback a random number in the range $[0,\delta]$ is applied to the probability $P_{io}$ that was used in the last update of the gate. In case of positive feedback the value is increased, in the case of negative feedback the value is decreased. The probabilities are limited to not exceed $0.99$ or drop below $0.01$. The rest of the probabilities are then normalized.

The effects of the feedback gate are immediately accessible to the MB. However, because MBs are networks, the signal that a feedback gate generates might need time to be relayed to the outputs via other gates. It is also possible that there is a delay between an agent's actions and the time it takes to receive new sensorial inputs that give a clue about the situation being improved or not. Thus, allowing feedback to occur only on the last action is not sufficient. Therefore, feedback gates can evolve the depth of a buffer that stores prior $P_{io}$ up to a depth of $4$ and when feedback is applied all the probabilities identified by the elements in the cue are altered. The $\delta$ is determined by evolution and can be different for each element in the cue.

\subsection*{Line of descent}
An agent was selected at random from the final generation to determine the line of descent (LOD) by tracing the ancestors to the first generation\cite{lenski2003evolutionary}. During this process the most recent common ancestor (MRCA) is quickly found. Observe that all mutations that swept the population can be found on the LOD, and the LOD contains all evolutionary changes that mattered.



\begin{thebibliography}{10}
\expandafter\ifx\csname url\endcsname\relax
  \def\url#1{\texttt{#1}}\fi
\expandafter\ifx\csname urlprefix\endcsname\relax\def\urlprefix{URL }\fi
\expandafter\ifx\csname doiprefix\endcsname\relax\def\doiprefix{DOI }\fi
\providecommand{\bibinfo}[2]{#2}
\providecommand{\eprint}[2][]{\url{#2}}

\bibitem{hebb2005organization}
\bibinfo{author}{Hebb, D.~O.}
\newblock \emph{\bibinfo{title}{The organization of behavior: A
  neuropsychological theory}} (\bibinfo{publisher}{Psychology Press},
  \bibinfo{year}{2005}).

\bibitem{baldwin1896new}
\bibinfo{author}{Baldwin, J.~M.}
\newblock \bibinfo{journal}{\bibinfo{title}{A new factor in evolution}}.
\newblock {\emph{\JournalTitle{The american naturalist}}}
  \textbf{\bibinfo{volume}{30}}, \bibinfo{pages}{441--451}
  (\bibinfo{year}{1896}).

\bibitem{sznajder2012adaptive}
\bibinfo{author}{Sznajder, B.}, \bibinfo{author}{Sabelis, M.} \&
  \bibinfo{author}{Egas, M.}
\newblock \bibinfo{journal}{\bibinfo{title}{How adaptive learning affects
  evolution: reviewing theory on the baldwin effect}}.
\newblock {\emph{\JournalTitle{Evolutionary biology}}}
  \textbf{\bibinfo{volume}{39}}, \bibinfo{pages}{301--310}
  (\bibinfo{year}{2012}).

\bibitem{hinton1987learning}
\bibinfo{author}{Hinton, G.~E.} \& \bibinfo{author}{Nowlan, S.~J.}
\newblock \bibinfo{journal}{\bibinfo{title}{How learning can guide evolution}}.
\newblock {\emph{\JournalTitle{Complex systems}}} \textbf{\bibinfo{volume}{1}},
  \bibinfo{pages}{495--502} (\bibinfo{year}{1987}).

\bibitem{fontanari1990effect}
\bibinfo{author}{Fontanari, J.} \& \bibinfo{author}{Meir, R.}
\newblock \bibinfo{journal}{\bibinfo{title}{The effect of learning on the
  evolution of asexual populations}}.
\newblock {\emph{\JournalTitle{Complex Systems}}} \textbf{\bibinfo{volume}{4}},
  \bibinfo{pages}{401--414} (\bibinfo{year}{1990}).

\bibitem{santos2015phenotypic}
\bibinfo{author}{Santos, M.}, \bibinfo{author}{Szathm{\'a}ry, E.} \&
  \bibinfo{author}{Fontanari, J.~F.}
\newblock \bibinfo{journal}{\bibinfo{title}{Phenotypic plasticity, the baldwin
  effect, and the speeding up of evolution: The computational roots of an
  illusion}}.
\newblock {\emph{\JournalTitle{Journal of theoretical biology}}}
  \textbf{\bibinfo{volume}{371}}, \bibinfo{pages}{127--136}
  (\bibinfo{year}{2015}).

\bibitem{dunlap2016reliability}
\bibinfo{author}{Dunlap, A.~S.} \& \bibinfo{author}{Stephens, D.~W.}
\newblock \bibinfo{journal}{\bibinfo{title}{Reliability, uncertainty, and costs
  in the evolution of animal learning}}.
\newblock {\emph{\JournalTitle{Current Opinion in Behavioral Sciences}}}
  \textbf{\bibinfo{volume}{12}}, \bibinfo{pages}{73--79}
  (\bibinfo{year}{2016}).

\bibitem{Ma:2014}
\bibinfo{author}{Ma, W.~J.}, \bibinfo{author}{Husain, M.} \&
  \bibinfo{author}{Bays, P.~M.}
\newblock \bibinfo{journal}{\bibinfo{title}{{Changing concepts of working
  memory}}}.
\newblock {\emph{\JournalTitle{Nature Neuroscience}}}
  \textbf{\bibinfo{volume}{17}}, \bibinfo{pages}{347--356}
  (\bibinfo{year}{2014}).

\bibitem{Nadel:2010}
\bibinfo{author}{Nadel, L.} \& \bibinfo{author}{Hardt, O.}
\newblock \bibinfo{journal}{\bibinfo{title}{{Update on Memory Systems and
  Processes}}}.
\newblock {\emph{\JournalTitle{Neuropsychopharmacology}}}
  \textbf{\bibinfo{volume}{36}}, \bibinfo{pages}{251--273}
  (\bibinfo{year}{2010}).

\bibitem{Kandel:2014}
\bibinfo{author}{Kandel, E.~R.}, \bibinfo{author}{Dudai, Y.} \&
  \bibinfo{author}{Mayford, M.~R.}
\newblock \bibinfo{journal}{\bibinfo{title}{{The Molecular and Systems Biology
  of Memory}}}.
\newblock {\emph{\JournalTitle{Cell}}} \textbf{\bibinfo{volume}{157}},
  \bibinfo{pages}{163--186} (\bibinfo{year}{2014}).

\bibitem{Squire:2011}
\bibinfo{author}{Squire, L.~R.} \& \bibinfo{author}{Wixted, J.~T.}
\newblock \bibinfo{journal}{\bibinfo{title}{{The Cognitive Neuroscience of
  Human Memory Since H.M.}}}
\newblock {\emph{\JournalTitle{Annual Review of Neuroscience}}}
  \textbf{\bibinfo{volume}{34}}, \bibinfo{pages}{259--288}
  (\bibinfo{year}{2011}).

\bibitem{Abraham:2005}
\bibinfo{author}{Abraham, W.~C.} \& \bibinfo{author}{Robins, A.}
\newblock \bibinfo{journal}{\bibinfo{title}{{Memory retention {\textendash} the
  synaptic stability versus plasticity dilemma}}}.
\newblock {\emph{\JournalTitle{Trends in Neurosciences}}}
  \textbf{\bibinfo{volume}{28}}, \bibinfo{pages}{73--78}
  (\bibinfo{year}{2005}).

\bibitem{McKenzie:2011}
\bibinfo{author}{McKenzie, S.} \& \bibinfo{author}{Eichenbaum, H.}
\newblock \bibinfo{journal}{\bibinfo{title}{{Consolidation and Reconsolidation:
  Two Lives of Memories?}}}
\newblock {\emph{\JournalTitle{Neuron}}} \textbf{\bibinfo{volume}{71}},
  \bibinfo{pages}{224--233} (\bibinfo{year}{2011}).

\bibitem{russell2009}
\bibinfo{author}{Russell, S.~J.} \& \bibinfo{author}{Norvig, P.}
\newblock \emph{\bibinfo{title}{Artificial intelligence: a modern approach (3rd
  edition)}} (\bibinfo{publisher}{Prentice Hall}, \bibinfo{year}{2009}).

\bibitem{schmidhuber1987evolutionary}
\bibinfo{author}{Schmidhuber, J.}
\newblock \bibinfo{journal}{\bibinfo{title}{Evolutionary principles in
  self-referential learning}}.
\newblock {\emph{\JournalTitle{On learning how to learn: The meta-meta-...
  hook.) Diploma thesis, Institut f. Informatik, Tech. Univ. Munich}}}
  (\bibinfo{year}{1987}).

\bibitem{watkins1989}
\bibinfo{author}{Watkins, C. J. C.~H.}
\newblock \emph{\bibinfo{title}{Learning from delayed rewards}}.
\newblock Ph.D. thesis, \bibinfo{school}{University of Cambridge England}
  (\bibinfo{year}{1989}).

\bibitem{mnih2013playing}
\bibinfo{author}{Mnih, V.} \emph{et~al.}
\newblock \bibinfo{journal}{\bibinfo{title}{Playing atari with deep
  reinforcement learning}}.
\newblock {\emph{\JournalTitle{arXiv preprint arXiv:1312.5602}}}
  (\bibinfo{year}{2013}).

\bibitem{Schmidhuber:2015}
\bibinfo{author}{Schmidhuber, J.}
\newblock \bibinfo{journal}{\bibinfo{title}{{Deep learning in neural networks:
  An overview}}}.
\newblock {\emph{\JournalTitle{Neural Networks}}}
  \textbf{\bibinfo{volume}{61}}, \bibinfo{pages}{85--117}
  (\bibinfo{year}{2015}).

\bibitem{Zhang:2012}
\bibinfo{author}{Zhang, H.}, \bibinfo{author}{Wu, W.} \& \bibinfo{author}{Yao,
  M.}
\newblock \bibinfo{journal}{\bibinfo{title}{{Boundedness and convergence of
  batch back-propagation algorithm with penalty for feedforward neural
  networks}}}.
\newblock {\emph{\JournalTitle{Neurocomputing}}} \textbf{\bibinfo{volume}{89}},
  \bibinfo{pages}{141--146} (\bibinfo{year}{2012}).

\bibitem{LeCun:2015}
\bibinfo{author}{LeCun, Y.}, \bibinfo{author}{Bengio, Y.} \&
  \bibinfo{author}{Hinton, G.}
\newblock \bibinfo{journal}{\bibinfo{title}{{Deep learning}}}.
\newblock {\emph{\JournalTitle{Nature}}} \textbf{\bibinfo{volume}{521}},
  \bibinfo{pages}{436--444} (\bibinfo{year}{2015}).

\bibitem{Kaleh:1994}
\bibinfo{author}{Kaleh, G.~K.} \& \bibinfo{author}{Vallet, R.}
\newblock \bibinfo{journal}{\bibinfo{title}{{Joint parameter estimation and
  symbol detection for linear or nonlinear unknown channels.}}}
\newblock {\emph{\JournalTitle{IEEE Trans. Communications ()}}}
  \textbf{\bibinfo{volume}{42}}, \bibinfo{pages}{2406--2413}
  (\bibinfo{year}{1994}).

\bibitem{Baggenstoss:2001}
\bibinfo{author}{Baggenstoss, P.~M.}
\newblock \bibinfo{journal}{\bibinfo{title}{{A modified Baum-Welch algorithm
  for hidden Markov models with multiple observation spaces}}}.
\newblock {\emph{\JournalTitle{IEEE Transactions on Speech and Audio
  Processing}}} \textbf{\bibinfo{volume}{9}}, \bibinfo{pages}{411--416}
  (\bibinfo{year}{2001}).

\bibitem{williams1992simple}
\bibinfo{author}{Williams, R.~J.}
\newblock \bibinfo{journal}{\bibinfo{title}{Simple statistical
  gradient-following algorithms for connectionist reinforcement learning}}.
\newblock {\emph{\JournalTitle{Machine learning}}}
  \textbf{\bibinfo{volume}{8}}, \bibinfo{pages}{229--256}
  (\bibinfo{year}{1992}).

\bibitem{kaelbling1996reinforcement}
\bibinfo{author}{Kaelbling, L.~P.}, \bibinfo{author}{Littman, M.~L.} \&
  \bibinfo{author}{Moore, A.~W.}
\newblock \bibinfo{journal}{\bibinfo{title}{Reinforcement learning: A survey}}.
\newblock {\emph{\JournalTitle{Journal of artificial intelligence research}}}
  \textbf{\bibinfo{volume}{4}}, \bibinfo{pages}{237--285}
  (\bibinfo{year}{1996}).

\bibitem{Arora:2012}
\bibinfo{author}{Arora, S.}, \bibinfo{author}{Hazan, E.} \&
  \bibinfo{author}{Kale, S.}
\newblock \bibinfo{journal}{\bibinfo{title}{{The Multiplicative Weights Update
  Method: a Meta-Algorithm and Applications.}}}
\newblock {\emph{\JournalTitle{Theory of Computing}}}  (\bibinfo{year}{2012}).

\bibitem{Freund:1999}
\bibinfo{author}{Freund, Y.} \& \bibinfo{author}{Schapire, R.~E.}
\newblock \bibinfo{journal}{\bibinfo{title}{{Adaptive game playing using
  multiplicative weights}}}.
\newblock {\emph{\JournalTitle{Games and Economic Behavior}}}
  \textbf{\bibinfo{volume}{29}}, \bibinfo{pages}{79--103}
  (\bibinfo{year}{1999}).

\bibitem{Marstaller:2013}
\bibinfo{author}{Marstaller, L.}, \bibinfo{author}{Hintze, A.} \&
  \bibinfo{author}{Adami, C.}
\newblock \bibinfo{journal}{\bibinfo{title}{{The Evolution of Representation in
  Simple Cognitive Networks}}}.
\newblock {\emph{\JournalTitle{Neural Computation}}}
  \textbf{\bibinfo{volume}{25}}, \bibinfo{pages}{2079--2107}
  (\bibinfo{year}{2013}).

\bibitem{kaelbling1998planning}
\bibinfo{author}{Kaelbling, L.~P.}, \bibinfo{author}{Littman, M.~L.} \&
  \bibinfo{author}{Cassandra, A.~R.}
\newblock \bibinfo{journal}{\bibinfo{title}{Planning and acting in partially
  observable stochastic domains}}.
\newblock {\emph{\JournalTitle{Artificial intelligence}}}
  \textbf{\bibinfo{volume}{101}}, \bibinfo{pages}{99--134}
  (\bibinfo{year}{1998}).

\bibitem{sims1994evolving}
\bibinfo{author}{Sims, K.}
\newblock \bibinfo{title}{Evolving virtual creatures}.
\newblock In \emph{\bibinfo{booktitle}{Proceedings of the 21st annual
  conference on Computer graphics and interactive techniques}},
  \bibinfo{pages}{15--22} (\bibinfo{organization}{ACM}, \bibinfo{year}{1994}).

\bibitem{stanley2002evolving}
\bibinfo{author}{Stanley, K.~O.} \& \bibinfo{author}{Miikkulainen, R.}
\newblock \bibinfo{journal}{\bibinfo{title}{Evolving neural networks through
  augmenting topologies}}.
\newblock {\emph{\JournalTitle{Evolutionary computation}}}
  \textbf{\bibinfo{volume}{10}}, \bibinfo{pages}{99--127}
  (\bibinfo{year}{2002}).

\bibitem{gauci2010autonomous}
\bibinfo{author}{Gauci, J.} \& \bibinfo{author}{Stanley, K.~O.}
\newblock \bibinfo{journal}{\bibinfo{title}{Autonomous evolution of topographic
  regularities in artificial neural networks}}.
\newblock {\emph{\JournalTitle{Neural computation}}}
  \textbf{\bibinfo{volume}{22}}, \bibinfo{pages}{1860--1898}
  (\bibinfo{year}{2010}).

\bibitem{spector2002genetic}
\bibinfo{author}{Spector, L.} \& \bibinfo{author}{Robinson, A.}
\newblock \bibinfo{journal}{\bibinfo{title}{Genetic programming and
  autoconstructive evolution with the push programming language}}.
\newblock {\emph{\JournalTitle{Genetic Programming and Evolvable Machines}}}
  \textbf{\bibinfo{volume}{3}}, \bibinfo{pages}{7--40} (\bibinfo{year}{2002}).

\bibitem{greve2015evolving}
\bibinfo{author}{Greve, R.~B.}, \bibinfo{author}{Jacobsen, E.~J.} \&
  \bibinfo{author}{Risi, S.}
\newblock \bibinfo{title}{Evolving neural turing machines}.
\newblock In \emph{\bibinfo{booktitle}{Neural Information Processing Systems:
  Reasoning, Attention, Memory Workshop}} (\bibinfo{year}{2015}).

\bibitem{yao1999evolving}
\bibinfo{author}{Yao, X.}
\newblock \bibinfo{journal}{\bibinfo{title}{Evolving artificial neural
  networks}}.
\newblock {\emph{\JournalTitle{Proceedings of the IEEE}}}
  \textbf{\bibinfo{volume}{87}}, \bibinfo{pages}{1423--1447}
  (\bibinfo{year}{1999}).

\bibitem{stanley2003evolving}
\bibinfo{author}{Stanley, K.~O.}, \bibinfo{author}{Bryant, B.~D.} \&
  \bibinfo{author}{Miikkulainen, R.}
\newblock \bibinfo{title}{Evolving adaptive neural networks with and without
  adaptive synapses}.
\newblock In \emph{\bibinfo{booktitle}{Evolutionary Computation, 2003. CEC'03.
  The 2003 Congress on}}, vol.~\bibinfo{volume}{4}, \bibinfo{pages}{2557--2564}
  (\bibinfo{organization}{IEEE}, \bibinfo{year}{2003}).

\bibitem{gomez2005evolving}
\bibinfo{author}{Gomez, F.} \& \bibinfo{author}{Schmidhuber, J.}
\newblock \bibinfo{journal}{\bibinfo{title}{Evolving modular fast-weight
  networks for control}}.
\newblock {\emph{\JournalTitle{Artificial Neural Networks: Formal Models and
  Their Applications--ICANN 2005}}} \bibinfo{pages}{750--750}
  (\bibinfo{year}{2005}).

\bibitem{urzelai2006evolution}
\bibinfo{author}{Urzelai, J.} \& \bibinfo{author}{Floreano, D.}
\newblock \bibinfo{journal}{\bibinfo{title}{Evolution of adaptive synapses:
  Robots with fast adaptive behavior in new environments}}.
\newblock {\emph{\JournalTitle{Evolution}}} \textbf{\bibinfo{volume}{9}}
  (\bibinfo{year}{2006}).

\bibitem{soltoggio2008evolutionary}
\bibinfo{author}{Soltoggio, A.}, \bibinfo{author}{Bullinaria, J.~A.},
  \bibinfo{author}{Mattiussi, C.}, \bibinfo{author}{D{\"u}rr, P.} \&
  \bibinfo{author}{Floreano, D.}
\newblock \bibinfo{title}{Evolutionary advantages of neuromodulated plasticity
  in dynamic, reward-based scenarios}.
\newblock In \emph{\bibinfo{booktitle}{Proceedings of the 11th International
  Conference on Artificial Life (Alife XI)}},
  \bibinfo{number}{LIS-CONF-2008-012}, \bibinfo{pages}{569--576}
  (\bibinfo{organization}{MIT Press}, \bibinfo{year}{2008}).

\bibitem{luderscontinual}
\bibinfo{author}{L{\"u}ders, B.}, \bibinfo{author}{Schl{\"a}ger, M.} \&
  \bibinfo{author}{Risi, S.}
\newblock \bibinfo{title}{Continual learning through evolvable neural turing
  machines}.
\newblock In \emph{\bibinfo{booktitle}{NIPS 2016 Workshop on Continual Learning
  and Deep Networks (CLDL 2016)}} (\bibinfo{year}{2016}).

\bibitem{tonelli2011relationships}
\bibinfo{author}{Tonelli, P.} \& \bibinfo{author}{Mouret, J.-B.}
\newblock \bibinfo{title}{On the relationships between synaptic plasticity and
  generative systems}.
\newblock In \emph{\bibinfo{booktitle}{Proceedings of the 13th annual
  conference on Genetic and evolutionary computation}},
  \bibinfo{pages}{1531--1538} (\bibinfo{organization}{ACM},
  \bibinfo{year}{2011}).

\bibitem{risi2012unified}
\bibinfo{author}{Risi, S.} \& \bibinfo{author}{Stanley, K.~O.}
\newblock \bibinfo{title}{A unified approach to evolving plasticity and neural
  geometry}.
\newblock In \emph{\bibinfo{booktitle}{Neural Networks (IJCNN), The 2012
  International Joint Conference on}}, \bibinfo{pages}{1--8}
  (\bibinfo{organization}{IEEE}, \bibinfo{year}{2012}).

\bibitem{coleman2012evolving}
\bibinfo{author}{Coleman, O.~J.} \& \bibinfo{author}{Blair, A.~D.}
\newblock \bibinfo{title}{Evolving plastic neural networks for online learning:
  review and future directions}.
\newblock In \emph{\bibinfo{booktitle}{Australasian Joint Conference on
  Artificial Intelligence}}, \bibinfo{pages}{326--337}
  (\bibinfo{organization}{Springer}, \bibinfo{year}{2012}).

\bibitem{greve2016evolving}
\bibinfo{author}{Greve, R.~B.}, \bibinfo{author}{Jacobsen, E.~J.} \&
  \bibinfo{author}{Risi, S.}
\newblock \bibinfo{title}{Evolving neural turing machines for reward-based
  learning}.
\newblock In \emph{\bibinfo{booktitle}{Proceedings of the 2016 on Genetic and
  Evolutionary Computation Conference}}, \bibinfo{pages}{117--124}
  (\bibinfo{organization}{ACM}, \bibinfo{year}{2016}).

\bibitem{Edlund:2011}
\bibinfo{author}{Edlund, J.~A.} \emph{et~al.}
\newblock \bibinfo{journal}{\bibinfo{title}{{Integrated Information Increases
  with Fitness in the Evolution of Animats}}}.
\newblock {\emph{\JournalTitle{PLoS Comput Biol}}}
  \textbf{\bibinfo{volume}{7}}, \bibinfo{pages}{e1002236}
  (\bibinfo{year}{2011}).

\bibitem{Olson:2013}
\bibinfo{author}{Olson, R.~S.}, \bibinfo{author}{Hintze, A.},
  \bibinfo{author}{Dyer, F.~C.}, \bibinfo{author}{Knoester, D.~B.} \&
  \bibinfo{author}{Adami, C.}
\newblock \bibinfo{journal}{\bibinfo{title}{{Predator confusion is sufficient
  to evolve swarming behaviour}}}.
\newblock {\emph{\JournalTitle{Journal of The Royal Society Interface}}}
  \textbf{\bibinfo{volume}{10}}, \bibinfo{pages}{20130305--20130305}
  (\bibinfo{year}{2013}).

\bibitem{Hintze:14}
\bibinfo{author}{Hintze, A.} \emph{et~al.}
\newblock \bibinfo{title}{{Evolution of Autonomous Hierarchy Formation and
  Maintenance}}.
\newblock In \emph{\bibinfo{booktitle}{Artificial Life 14: Proceedings of the
  Fourteenth International Conference on the Synthesis and Simulation of Living
  Systems}}, \bibinfo{pages}{366--367} (\bibinfo{publisher}{The MIT Press},
  \bibinfo{year}{2014}).

\bibitem{Joshi:2013}
\bibinfo{author}{Joshi, N.~J.}, \bibinfo{author}{Tononi, G.} \&
  \bibinfo{author}{Koch, C.}
\newblock \bibinfo{journal}{\bibinfo{title}{{The minimal complexity of adapting
  agents increases with fitness}}}.
\newblock {\emph{\JournalTitle{PLoS Comput Biol}}}  (\bibinfo{year}{2013}).

\bibitem{Albantakis:2014}
\bibinfo{author}{Albantakis, L.}, \bibinfo{author}{Hintze, A.},
  \bibinfo{author}{Koch, C.}, \bibinfo{author}{Adami, C.} \&
  \bibinfo{author}{Tononi, G.}
\newblock \bibinfo{journal}{\bibinfo{title}{{Evolution of Integrated Causal
  Structures in Animats Exposed to Environments of Increasing Complexity}}}.
\newblock {\emph{\JournalTitle{PLoS Comput Biol}}}
  \textbf{\bibinfo{volume}{10}}, \bibinfo{pages}{e1003966--19}
  (\bibinfo{year}{2014}).

\bibitem{Schossau2016}
\bibinfo{author}{Schossau, J.}, \bibinfo{author}{Adami, C.} \&
  \bibinfo{author}{Hintze, A.}
\newblock \bibinfo{journal}{\bibinfo{title}{{Information-Theoretic
  Neuro-Correlates Boost Evolution of Cognitive Systems}}}.
\newblock {\emph{\JournalTitle{Entropy}}} \textbf{\bibinfo{volume}{18}},
  \bibinfo{pages}{6--22} (\bibinfo{year}{2016}).

\bibitem{kvam2015}
\bibinfo{author}{Kvam, P.}, \bibinfo{author}{Cesario, J.},
  \bibinfo{author}{Schossau, J.}, \bibinfo{author}{Eisthen, H.} \&
  \bibinfo{author}{Hintze, A.}
\newblock \bibinfo{journal}{\bibinfo{title}{Computational evolution of
  decision-making strategies}}.
\newblock {\emph{\JournalTitle{arXiv preprint arXiv:1509.05646}}}
  (\bibinfo{year}{2015}).

\bibitem{Kvam2017}
\bibinfo{author}{Kvam, P.} \& \bibinfo{author}{Arend, H.}
\newblock \bibinfo{journal}{\bibinfo{title}{Rewards, risks, and reaching the
  right strategy: Evolutionary paths from heuristics to optimal decisions}}.
\newblock {\emph{\JournalTitle{Evolutionary Behavioral Sciences, invited
  submission for the Special Issue on Studying Evolved Cognitive Mechanisms}}}
  (\bibinfo{year}{(under review)}).

\bibitem{Chapman:2013}
\bibinfo{author}{Chapman, S.}, \bibinfo{author}{Knoester, D.~B.},
  \bibinfo{author}{Hintze, A.} \& \bibinfo{author}{Adami, C.}
\newblock \bibinfo{journal}{\bibinfo{title}{{Evolution of an artificial visual
  cortex for image recognition.}}}
\newblock {\emph{\JournalTitle{ECAL}}} \bibinfo{pages}{1067--1074}
  (\bibinfo{year}{2013}).

\bibitem{chapman2017evolution}
\bibinfo{author}{Chapman, S.~D.}, \bibinfo{author}{Adami, C.},
  \bibinfo{author}{Wilke, C.~O.} \& \bibinfo{author}{KC, D.~B.}
\newblock \bibinfo{journal}{\bibinfo{title}{The evolution of logic circuits for
  the purpose of protein contact map prediction}}.
\newblock {\emph{\JournalTitle{PeerJ}}} \textbf{\bibinfo{volume}{5}},
  \bibinfo{pages}{e3139} (\bibinfo{year}{2017}).

\bibitem{russell2005ai}
\bibinfo{author}{Russell, S.} \& \bibinfo{author}{Norvig, P.}
\newblock \bibinfo{journal}{\bibinfo{title}{Ai a modern approach}}.
\newblock {\emph{\JournalTitle{Learning}}} \textbf{\bibinfo{volume}{2}},
  \bibinfo{pages}{4} (\bibinfo{year}{2005}).

\bibitem{clark1998being}
\bibinfo{author}{Clark, A.}
\newblock \emph{\bibinfo{title}{Being there: Putting brain, body, and world
  together again}} (\bibinfo{publisher}{MIT press}, \bibinfo{year}{1998}).

\bibitem{french1999catastrophic}
\bibinfo{author}{French, R.~M.}
\newblock \bibinfo{journal}{\bibinfo{title}{Catastrophic forgetting in
  connectionist networks}}.
\newblock {\emph{\JournalTitle{Trends in cognitive sciences}}}
  \textbf{\bibinfo{volume}{3}}, \bibinfo{pages}{128--135}
  (\bibinfo{year}{1999}).

\bibitem{ellefsen2015neural}
\bibinfo{author}{Ellefsen, K.~O.}, \bibinfo{author}{Mouret, J.-B.} \&
  \bibinfo{author}{Clune, J.}
\newblock \bibinfo{journal}{\bibinfo{title}{Neural modularity helps organisms
  evolve to learn new skills without forgetting old skills}}.
\newblock {\emph{\JournalTitle{PLoS Comput Biol}}}
  \textbf{\bibinfo{volume}{11}}, \bibinfo{pages}{e1004128}
  (\bibinfo{year}{2015}).

\bibitem{ciregan2012multi}
\bibinfo{author}{Ciregan, D.}, \bibinfo{author}{Meier, U.} \&
  \bibinfo{author}{Schmidhuber, J.}
\newblock \bibinfo{title}{Multi-column deep neural networks for image
  classification}.
\newblock In \emph{\bibinfo{booktitle}{Computer Vision and Pattern Recognition
  (CVPR), 2012 IEEE Conference on}}, \bibinfo{pages}{3642--3649}
  (\bibinfo{organization}{IEEE}, \bibinfo{year}{2012}).

\bibitem{silver2016mastering}
\bibinfo{author}{Silver, D.} \emph{et~al.}
\newblock \bibinfo{journal}{\bibinfo{title}{Mastering the game of go with deep
  neural networks and tree search}}.
\newblock {\emph{\JournalTitle{Nature}}} \textbf{\bibinfo{volume}{529}},
  \bibinfo{pages}{484--489} (\bibinfo{year}{2016}).

\bibitem{MABE2016}
\bibinfo{author}{Hintze, A.} \& \bibinfo{author}{Bohm, C.}
\newblock \bibinfo{title}{Mabe}.
\newblock \bibinfo{howpublished}{\url{https://github.com/ahnt/MABE}}
  (\bibinfo{year}{2016}).

\bibitem{blickle1996comparison}
\bibinfo{author}{Blickle, T.} \& \bibinfo{author}{Thiele, L.}
\newblock \bibinfo{journal}{\bibinfo{title}{A comparison of selection schemes
  used in evolutionary algorithms}}.
\newblock {\emph{\JournalTitle{Evolutionary Computation}}}
  \textbf{\bibinfo{volume}{4}}, \bibinfo{pages}{361--394}
  (\bibinfo{year}{1996}).

\bibitem{lenski2003evolutionary}
\bibinfo{author}{Lenski, R.~E.}, \bibinfo{author}{Ofria, C.},
  \bibinfo{author}{Pennock, R.~T.} \& \bibinfo{author}{Adami, C.}
\newblock \bibinfo{journal}{\bibinfo{title}{The evolutionary origin of complex
  features}}.
\newblock {\emph{\JournalTitle{Nature}}} \textbf{\bibinfo{volume}{423}},
  \bibinfo{pages}{139--144} (\bibinfo{year}{2003}).

\end{thebibliography}


\section*{Acknowledgements (not compulsory)}
This work is in part supported by the NSF BEACON-center for the study of evolution in action under Cooperative Agreement DBI-0939454. We would like to thank Christoph Adami, Charles C. Ofria, and Fred C. Dyer for useful discussions.

\section*{Author contributions statement}

L.S. and A.H. conceived of the experiments and wrote the manuscript. L.S. performed all computational experiments and data analysis.

\section*{Additional information}
The authors declare no competing financial interests


\end{document}